\documentclass[pdflatex,sn-mathphys-num]{sn-jnl}

\usepackage{graphicx}%
\usepackage{multirow}%
\usepackage{amsmath,amssymb,amsfonts}%
\usepackage{amsthm}%
\usepackage{mathrsfs}%
\usepackage[title]{appendix}%
\usepackage{xcolor}%
\usepackage{textcomp}%
\usepackage{manyfoot}%
\usepackage{booktabs}%
\usepackage{algpseudocode}%
\usepackage{algorithm}%
\usepackage{algorithmicx}%
\usepackage{algpseudocode}%
\usepackage{listings}%
\usepackage{caption}%
\usepackage{subcaption}%

\newboolean{showcomments}
\setboolean{showcomments}{true}
\ifthenelse{\boolean{showcomments}}
{\newcommand{\nb}[2]{
\fbox{\bfseries\sffamily\scriptsize#1}
{\sf\small$\blacktriangleright$\textit{#2}$\blacktriangleleft$}
}
}
{\newcommand{\nb}[2]{}
}

\algnewcommand\algorithmicforeach{\textbf{for each}}
\algdef{S}[FOR]{ForEach}[1]{\algorithmicforeach\ #1\ \algorithmicdo}

\theoremstyle{thmstyleone}%
\theoremstyle{thmstyletwo}%

\theoremstyle{thmstylethree}%

\begin{document}

\title[Article Title]{Leveraging Time Series Categorization and Temporal Fusion Transformers to Improve Cryptocurrency Price Forecasting}








\author[1]{\fnm{Arash} \sur{Peik}}\email{peik@stu.yazd.ac.ir}

\author*[1]{\fnm{Mohammad Ali} \sur{Zare Chahooki}}\email{chahooki@yazd.ac.ir}

\author[2]{\fnm{Amin} \sur{Milani Fard }}\email{amilanif@nyit.edu}

\author[1]{\fnm{Mehdi} \sur{Agha Sarram}}\email{mehdi.sarram@yazd.ac.ir}

\affil*[1]{\orgdiv{Department of Computer Engineering}, \orgname{Yazd University}, \orgaddress{\city{Yazd}, \postcode{89158-18411}, \country{Iran}}}

\affil[2]{\orgdiv{Department of Computer Science}, \orgname{New York Institute of Technology}, \orgaddress{\city{Vancouver}, \postcode{BC V5M 4X3}, \country{Canada}}}


\abstract{Organizing and managing cryptocurrency portfolios and decision-making on transactions is crucial in this market. Optimal selection of assets is one of the main challenges that requires accurate prediction of the price of cryptocurrencies. In this work, we categorize the financial time series into several similar subseries to increase prediction accuracy by learning each subseries category with similar behavior. For each category of the subseries, we create a deep learning model based on the attention mechanism to predict the next step of each subseries. Due to the limited amount of cryptocurrency data for training models, if the number of categories increases, the amount of training data for each model will decrease, and some complex models will not be trained well due to the large number of parameters. To overcome this challenge, we propose to combine the time series data of other cryptocurrencies to increase the amount of data for each category, hence increasing the accuracy of the models corresponding to each category.}

\keywords{Cryptocurrency trading, Financial time series, Temporal Fusion Transformers, Categorization of time series}



\maketitle

\section{Introduction}\label{sec1}

 
Cryptocurrencies market price fluctuations analysis and prediction are among the most popular topics in the cryptocurrencies research field \cite{corbet2019cryptocurrencies}. Digital currency fluctuations depend on factors such as volume, the difficulty of cryptocurrency mining, popularity, alternative coin prices, transaction costs, social media sentiments, and countries' laws and regulations \citep{angela2020factors,sovbetov2018factors}, and hence, price forecasting is a challenging task. 

A financial time series is usually a set of discrete sampled values for the price of a commodity or currency. The chosen time intervals between consecutive price values can differ significantly from seconds and minutes to days and months. Predicting financial time series is a critical task in financial markets, where machine learning techniques have shown to be effective. One prevalent approach is using supervised learning, such as regression and classification, which leverage historical data to identify patterns and relationships that can be used to forecast future values \cite{yamin2023cryptocurrency}. Akyildirim et al. \cite{akyildirim2023forecasting} examined the performance of different machine learning algorithms for predicting the mid-price movement for Bitcoin. Chen et al. \cite{chen2019predicting} used classic machine learning to predict Ethereum price. Fazlollahi and Ebrahimijam \cite{fazlollahi2023predicting} presented a multi-input Long Short Term Memory (LSTM) neural network with technical analysis indicator lags as input to forecast the price returns of Bitcoin, Ethereum, and Ripple. Bouteska et al. \cite{bouteska2024cryptocurrency} provided a comparative analysis of ensemble learning and deep learning models for cryptocurrencies price prediction. Feature engineering plays a crucial role in enhancing the performance of such models as selecting relevant financial indicators and appropriately transforming them can significantly improve the prediction accuracy.

\textbf{Motivation.} The Temporal Fusion Transformer (TFT) \cite{lim2021temporal} is a recent sophisticated attention-based deep neural network architecture for forecasting time series as an extension of the traditional Transformer architecture \cite{vaswani2017attention}. TFT incorporates temporal fusion mechanisms to capture temporal patterns and dependencies within sequential data. This model has shown state-of-the-art performance in various time series forecasting benchmarks and can effectively capture the intricate temporal patterns in financial markets \cite{hu2021stock}. Given the dynamic nature of financial data influenced by external factors, the TFT's adaptability and capability to model complex dependencies make it a promising tool for accurate and robust financial time series predictions. Considering the functionality of the TFT model and its accuracy in repeatable processes and events over time, instead of directly applying it on a given time series, in this work we investigate the effect of splitting the input time series into similar subseries (categories) and feeding these subseries as separate time series to TFT so that the model learns the repeated behavior of each category. Our hypothesis is that a more accurate prediction of subseries could result in higher accuracy of the series prediction.

\textbf{Contributions.} In this work we propose to categorize the financial time series into several subseries with similar behavior and train a TFT model for each category of the subseries separately. We train these models to predict subseries of other cryptocurrencies that behave similar to the corresponding category. By using different selectors for each cryptocurrency, which recognize the subseries category, it would be possible to predict each cryptocurrency's bullish or bearish status. By training repeated events in the time series, we increased the prediction accuracy of the model compared to deep models such as LSTM and simple TFT, and increased the profitability of algorithmic trading.

\textbf{Organization.} The rest of this paper is as follows. Section 2 reviews the previous related work. Section 3 describes our proposed approach. Section 4 provides the results of our experiments. Section 5 presents discussion. Finally, Section 6 draws conclusions and suggests future works.

\section{Related Work}\label{subsec2}

Kumar et al. \cite{kumar2018comparative} researched stock market behavior through price history and examined machine learning algorithms. 
Random Forest was shown to give the best results for large data sets while Naive Bayesian provided the highest accuracy for small datasets. Song and Lee \cite{song2019design} applied deep learning and neural networks to determine input features that can be used to predict stock values optimally. They observed that out of a large set of input features, only a few of them such as RSI, stochastic indicator, and CCI indicator affect the stock price. 

The authors in \cite{werawithayaset2019stock} applied methods such as Multilayer Perceptron and Partial Least Squares to predict stock price fluctuations in the Thai stock exchange. While their focus was not primarily on long-term investment decisions, they provided evidence that the Partial Least Squares method, accompanied by minimum error, outperformed Sequential Minimal Optimization and Multilayer Perceptron in minimizing error for the specific dataset chosen.

Xingzhou et al. \cite{xingzhou2019predictive} focused on the influence of indices in stock price prediction with the help of deep learning. This model identifies the variables and the relationship between the indices, overcomes the limitations of the traditional linear model, and uses LSTM to understand the dynamics of the S\&P 500 index (a list of the top 500 stocks in the New York Stock Exchange). 
However, the study has limitations; the difference between the predicted value and the actual value becomes large after a certain point and, therefore, cannot be used to develop a system to provide a profitable trading strategy.


Sarode et al. \cite{sarode2019stock} hypothesized that investors' emotions and feelings affect their reactions to specific news. They suggest a system that advises buyers to buy stocks by combining historical and real-time data and news analysis with LSTM for forecasting. In the RNN model, the first layer takes the latest trading data and technical indicators as input, then LSTM, a compressed layer, and finally, the output layer gives the predicted value. These predicted values are then joined with the summarized data from the news analytics to show a report with the percentage change.

Researchers sought to investigate how users' emotions affect prices \cite{ingle2016hidden,singh2019stock}. Ingle and Deshmukh used a hidden Markov model in their research \cite{ingle2016hidden}, but solely to detect users' opinions. They computed TF-IDF to calculate word count scores from news channels and subsequently generated an HMM model. While this model provided relatively accurate predictions for the stock market, it could not match this accuracy level for the cryptocurrency market. This is because, in the cryptocurrency market, large transfers outside of exchanges can instantaneously alter the direction of cryptocurrency value growth and have a higher impact than published content. Singh et al. \cite{singh2019stock} evaluated different research methods and found that some models work better with historical data and others with sentiment data. They observed that not all the approaches they reviewed used sentiment data to build their prediction model, although historical data analysis was necessary in each case to predict stock prices. They found that models that incorporated both historical and sentiment data were more accurate than those that used only one type of data. The authors also noted that the best performing models were those that used a combination of different machine learning algorithms. 



Ramos-P{\'e}rez et al. \cite{ramos2021multi} used multiple transformers to predict the stock price of the S\&P market. Such models have already been successfully used in natural language processing. This article also adapts the traditional transformer layers for use in fluctuation prediction models. The experimental results obtained in their work show that the hybrid models based on multi-transformer layers are more accurate.

Anbaee Farimani et al. \cite{farimani2021leveraging} consider the conceptual relationship between news documents as well as news sentiment and technical indicators to predict the market. They propose a concept-based news representation learning method to create a low-dimensional space of latent economic concepts. They later presented a model by leveraging sentiment mood time series via the probability distribution of news embedding generated through a BERT-based transformer fine-tuned for financial domain sentiment analysis and informative market data \cite{farimani2022investigating}. The authors recently presented an adaptive behaviour model for price regression \cite{farimani2024adaptive} built on top of their previous works. It extracts news-based features via a recurrent temporal convolution layer, and market and mood-based features with two separate recurrent layers. The model applies an adaptive fusion strategy that learns how to balance multimode of extracted features from various information sources to handle multimodalities of information sources.

\section{The Proposed Approach}\label{sec2}

The main idea of our work is to to increase the prediction accuracy by categorizing the financial time series into several similar subseries and more accurate learning of classes of similar subseries. In this method, we use a Markov chain to predict the next subseries category after finding the category of the last observed subseries. Finally, with the help of a transformer-based model related to that category, we predict the value of the cryptocurrency at the next moment.


\begin{figure}
    \centering
    \includegraphics[trim={0.95cm 13.5cm 33cm 5cm}, width=0.98\linewidth]{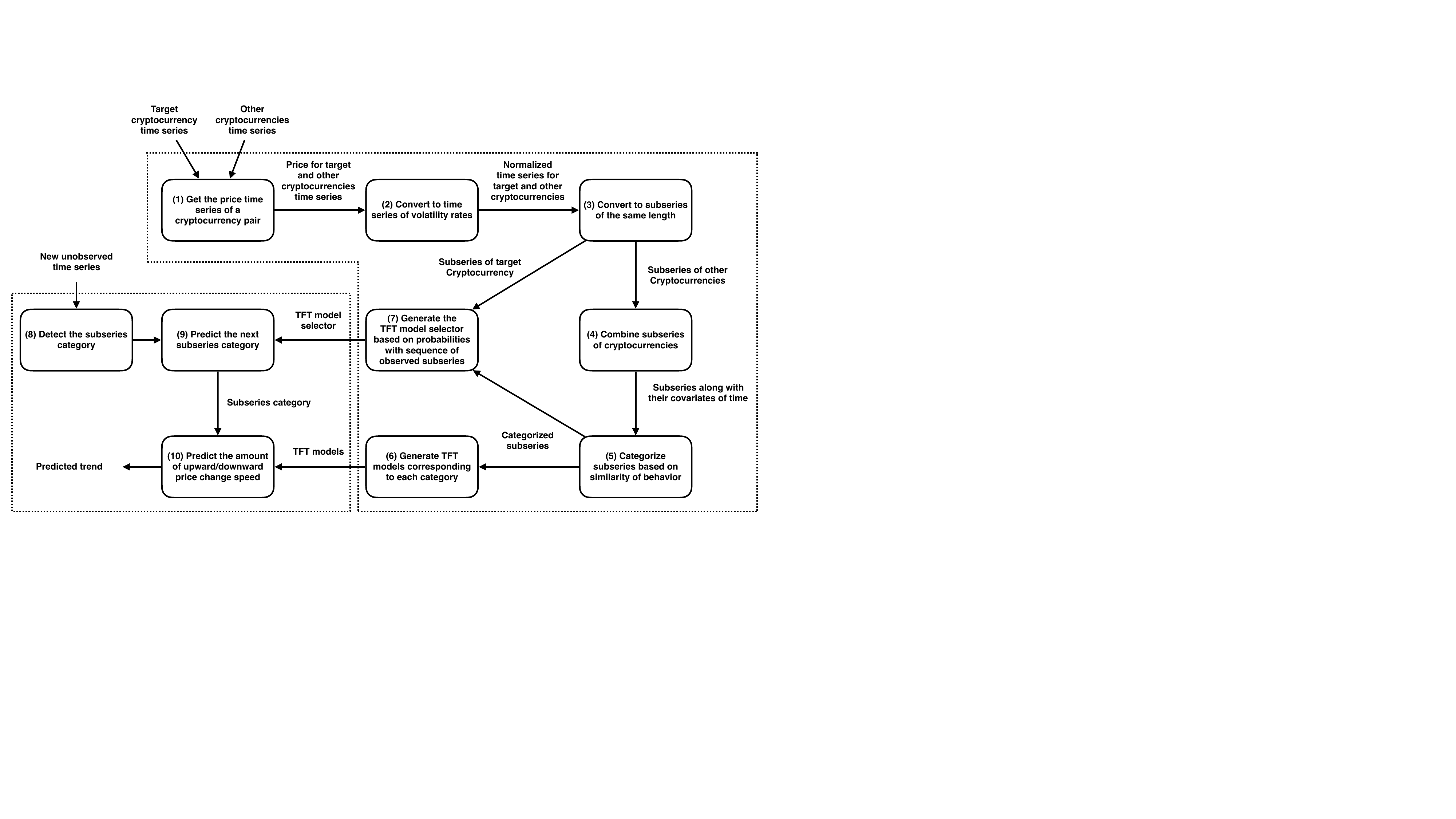}
    \caption{Overview of our proposed method.}
    \label{fig:mymodel}
\end{figure}

Figure \ref{fig:mymodel} depicts the overview of our proposed approach. We extract the Open, High, Low, Close, and Volume (OHLCV) information from a target cryptocurrency time series (block 1). Then, because a cryptocurrency may fluctuate in completely different price ranges over time, and to use the entire time series in model training, we change the price time series to the volatility rate time series (block 2). For example, on October 13, 2023, the price of Bitcoin was around \$26,750. Several days before and after, the price has increased and decreased in the same range, and our goal is to predict this increase and decrease in prices. Five months later, the price of Bitcoin fluctuated around \$71,500. If we use the price time series, the decrease in the price of \$100 will be considered the same in the entire time series if the decrease in the price of \$100 in October 2023 is much more significant than this decrease in March 2024. By changing the time series of the price to the time series of the volatility, in short time frames of a few minutes, we have somehow normalized the financial values in the range of -1 to 1 (percent). Because the rise or fall of the cryptocurrency price and its amount are important to us, this change makes the model more focused on this issue.

In the next step, we extract subseries with the same length from the created time series (block 3). These subseries are created by moving a sliding window on the time series. We will do all the steps of block 1 to 3 for the time series of other cryptocurrencies as well. We then combined similar series of other cryptocurrencies with the data (block 4) to increase the training data for better learning of the models. Next, we categorized them based on the upward or downward volatility of the price of each time frame compared to the previous one (block 5). We explain this categorization in more detail in section \ref{subsec2cat}. 

We then create a TFT model for each category of subseries (block 6). From the subseries of target cryptocurrency extracted from block 3 and their corresponding categories, we create a probabilistic model to predict the next category after observing a category of subseries (block 7). This model predicts upward or downward volatility of the price of the next time frame compared to the current time frame. With this prediction, we will not know the exact amount of upward or downward volatility, which will be predicted in the next step. We note that the volatility may be decreasing, but the price is increasing, or the volatility may be increasing, but the price is decreasing, which we will explain further. When a new subseries is observed until time $t$, we intend to predict the price and status in time $t$+1. After detecting the observed subseries category (block 8), we send it to a model selector to predict the next subseries category with probabilities (block 9). Finally, we send the subseries to the TFT model corresponding to the predicted category to predict for $t$+1 (block 10).

Algorithm \ref{alg1} shows the steps of our proposed method. It gets as input the time series of all available cryptocurrencies for training the TFT models. The preprocessing steps (lines 1-5) include conversion of price time series to price volatility time series (line 2 corresponding to block 2 in Figure \ref{fig:mymodel}), converting the time series into subseries based on the movement of the sliding window to the length of the number of time frames $n$ (line 3 according to block3). We considered $n$ to be 8 so that the trained model can predict the value of the $8^{th}$ time frame after seeing a time series including 7 time frames. Next, all subseries of cryptocurrencies are merged together in line 4 corresponding to block 4. Lines 6-9 perform subseries categorization (block 5) and build $allCatSubseries$ that is a key-value array, where the key is $catID$ and the value is a list of subsets key is the $catID$ and value is a list of subseries. TFT models will be trained in lines 10-12, and finally, in lines 13-17 according to the desired cryptocurrency for prediction, the selective model specific to the same cryptocurrency will be produced according to the categories after performing the aforementioned preprocessing steps.

\begin{algorithm}
\caption{TFT Model and Model Selector Generator.}
\label{alg1}
\begin{algorithmic}[1]

\Require Number of time frames $n$, cryptocurrencies time series $T$, cryptocurrency to predict $c$ 

\Ensure TFT model $M$ for all cryptocurrencies, and TFT model selector $S$ for cryptocurrency $c$ 

\ForEach {cryptocurrency time series $t \in T$}
        \State $v \leftarrow PriceToVolatility(t)$
        \State $subseries \leftarrow SplitToSubseries(v, n)$ 
        \State $allSubseries \leftarrow allSubseries \cup subseries $
\EndFor
\ForEach {$sub \in allSubseries $}
        \State $catID \leftarrow Categorize(sub)$    
        \State $allCatSubseries[catID] \leftarrow allCatSubseries[catID] \cup {sub} $
\EndFor

\ForEach {key-value \{$catID => catSubseries\} \in allCatSubseries $} 
        \State $M[catID] \leftarrow LearnTFTModel(catSubseries)$     
\EndFor

\ForEach {$cryptoTS \in T[c]$}
        \State $v \leftarrow PriceToVolatility(cryptoTS)$
        \State $subseries \leftarrow SplitToSubseries(v, n)$ 
        \State $S \leftarrow LearnTFTSelectorModel(allCatSubseries,subseries) $
\EndFor

\State \Return \{$M$,$S$\}

\end{algorithmic}
\end{algorithm}








\subsection{Data Preprocessing}\label{subsec2}

We collected a dataset of Binance transactions in 1-minute time frames for 2 years, which has not been published before at this granularity. The details of the dataset are described in section \ref{datasetsubsec2}. The important features used are "volume", "closing price", "number of trades", "quote Asset Volume". Among the price features, we will change the price of the close moment to the feature of the percentage change of the price compared to the previous moment as Volatility rate. For example, consider a subseries of Litecoin price based on Tether as [83.97 , 83.92 , 83.79 , 83.67 , 83.67 , 83.64 , 83.61 , 83.61]. We change this subseries to Volatility rate subseries using $V_t =  (\frac{P_t}{P_{t-1}}-1) *100$, that modifies time series to [-0.05 , -0.15 , -0.14 , 0 , -0.03 , -0.03 , 0]. Note that since $P_{t-1}$ is used to produce the subseries of changes and the first value of the subseries does not have the previous moment, the number of members of the changed subseries is one less, i.e. 7 numbers compared to 8 in this example.


This will create a new feature based on the price, which does not depend on the type of cryptocurrency and its price change interval. In order to better train models that need large amounts of data for training, the data of different cryptocurrencies can be used collectively to improve the quality of training.

To normalize some other features, such as the "number of trades", we divide them by the maximum value of that feature during a period when the market is neither strongly bullish nor strongly bearish. If the market is bullish or bearish and the value of these features is greater than the selected maximum, we consider it as 1 in normalization. Considering that we have set the closing price as a prediction criterion, we replace the open, high and low features for each period with their ratio to the closing price of that period. In other words, we perform feature extraction by dividing the values of each attribute by the value of the "close" attribute of the previous record.

In a deep model based on transformers, if the mentioned features and indicators, such as Moving Average Convergence/Divergence, are used, it can effectively improve the accuracy, which is also mentioned in the results section of this research. However, after checking the correlation of the features, we realized that the current close price and the features extracted from it do not have a significant correlation with other features. Hence, we only used the "close price volatility rate" feature for the sake of simplicity of the models. Moreover, we entered the position of the moment in the time series as covariates in the models.

\subsection{Time Series Categorization}\label{subsec2cat}

\begin{figure}
	\centering
	\begin{subfigure}{0.27\linewidth}
		\includegraphics[width=0.9\linewidth]{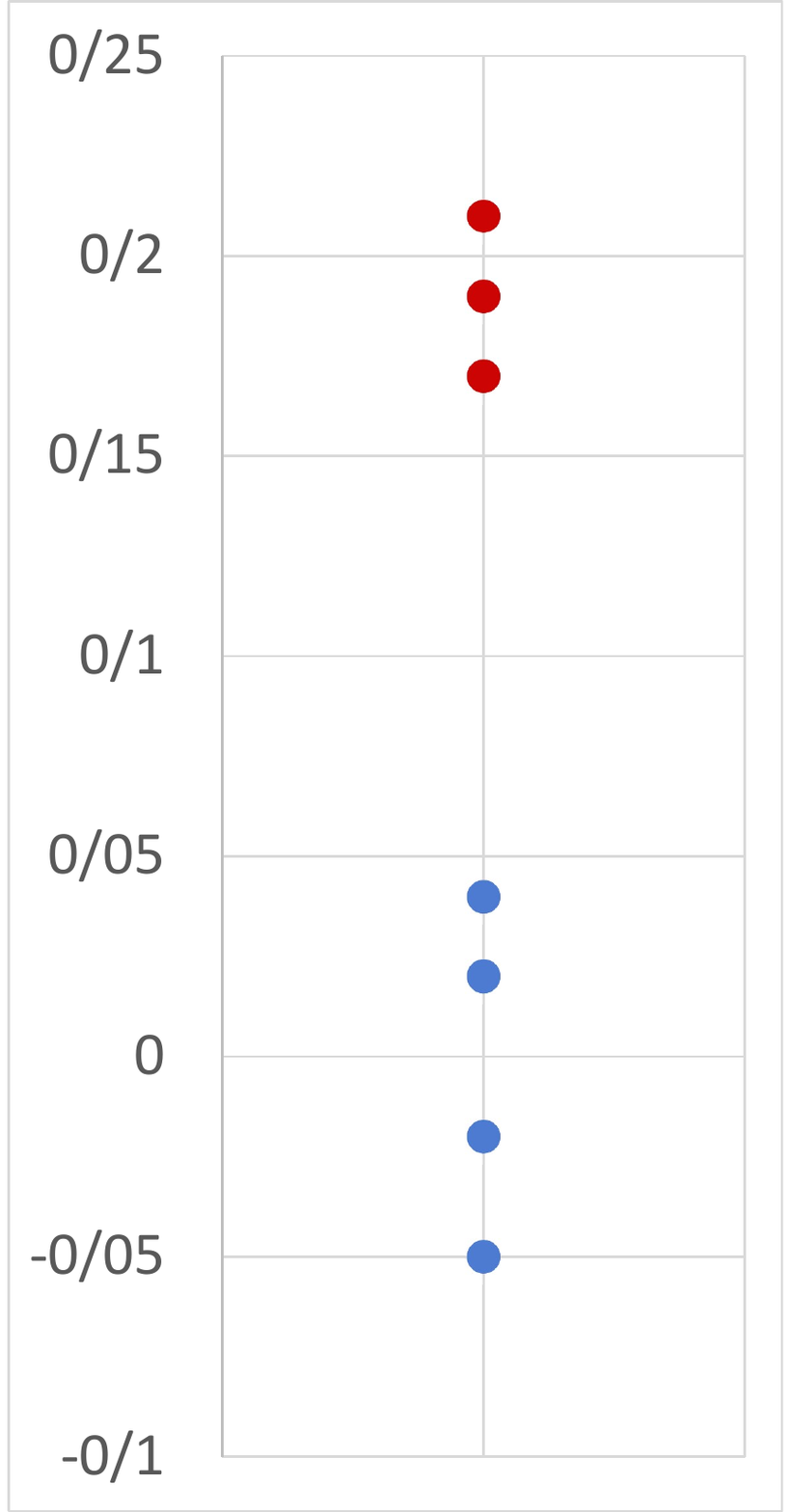}
		\caption{Cluster with different behavior}
		\label{fig:subfigAA}
	\end{subfigure}
	\begin{subfigure}{0.72\linewidth}
		\includegraphics[width=\linewidth]{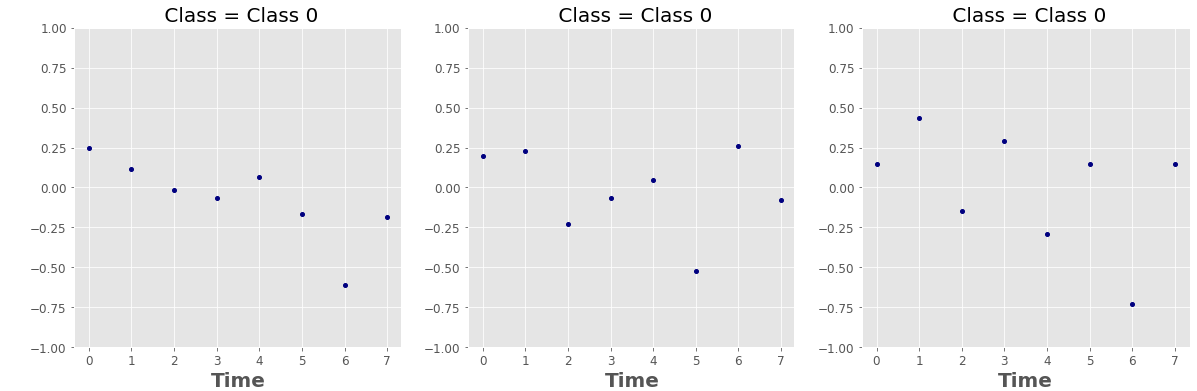}
		\caption{Subseries with different behavior in a cluster}
		\label{fig:subfigBB}
	\end{subfigure}
	
	\caption{Cluster with different behavior.}
	\label{fig:Cluster}
\end{figure}

Clustering a financial time series does not generally divide the subseries into behavioral similar clusters. 
Figure \ref{fig:subfigAA} shows an example of clustering in 1-dimensional space, which shows the red points of one cluster and the blue points of another cluster. Blue cluster points are different from each other in terms of behavior. If the chart shows the volatility of the price, two points of this cluster show an increase in the price and the other two points show a decrease in the price. Figure \ref{fig:subfigBB} also shows the three subseries that are placed in one cluster after running the $k$-means algorithm to divide into 256 clusters. These three subseries have completely different behaviors compared to each other.

Clustering in the $n$-dimensional space, where $n$ equals the number of moments of each subseries, results in placing subseries with similar vectors in the same category. Thus, two subseries that differ in behavior but have small distance in the $n$-dimensional space may be put in the same cluster. It is also possible that two similar subseries in behavior and logic may not be placed in the same cluster. On the other hand, the clustering operation requires its own processing time and resources. For these reasons, they can be labeled with little processing load without using clustering and considering the behavior of each subseries. Since in the labeling operation, the categories are already known and do not require special learning, we use the term "categorization" for this.

\begin{figure}
    \centering
    \includegraphics[width=0.7\linewidth]{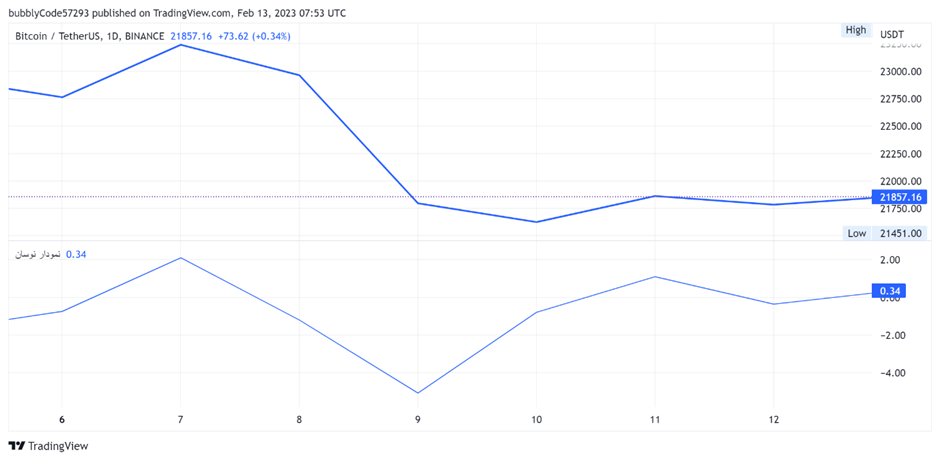}
    \caption{Time subseries of 7 consecutive moments of the Bitcoin-Tether currency pair}
    \label{fig:time-subseries-7}
\end{figure}

This Categorization is such that if we divide the initial time series into subseries with 8 consecutive moments and consider the behaviors to be only two states of decreasing or increasing compared to the previous moment, each subseries can be a member of \(2^7\) different states (categories). Figure \ref{fig:time-subseries-7} (upper part) shows a subseries consisting of 7 consecutive moments of the time series of the price of Bitcoin against Tether. The bottom part of figure \ref{fig:time-subseries-7} shows the price volatility of the top part.

\begin{figure}
    \centering
    \includegraphics[width=0.7\linewidth]{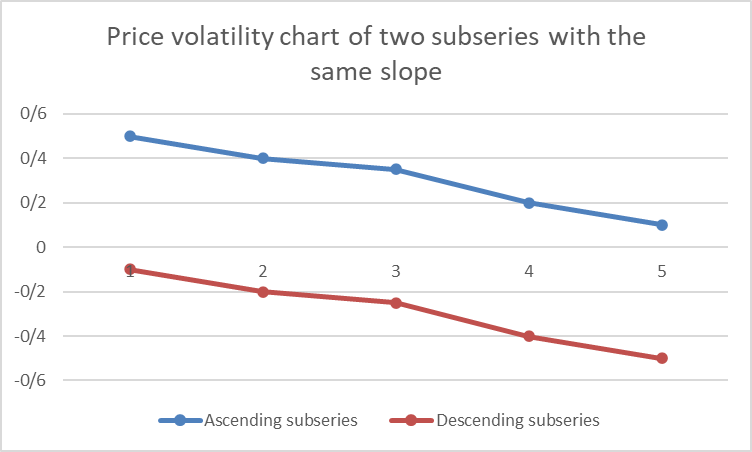}
    \caption{Two subseries with similar appearance, same acceleration, but different behavior.}
    \label{fig:enter-label}
\end{figure}

\begin{figure}
	\centering
	\begin{subfigure}{0.45\linewidth}
		\includegraphics[width=\linewidth]{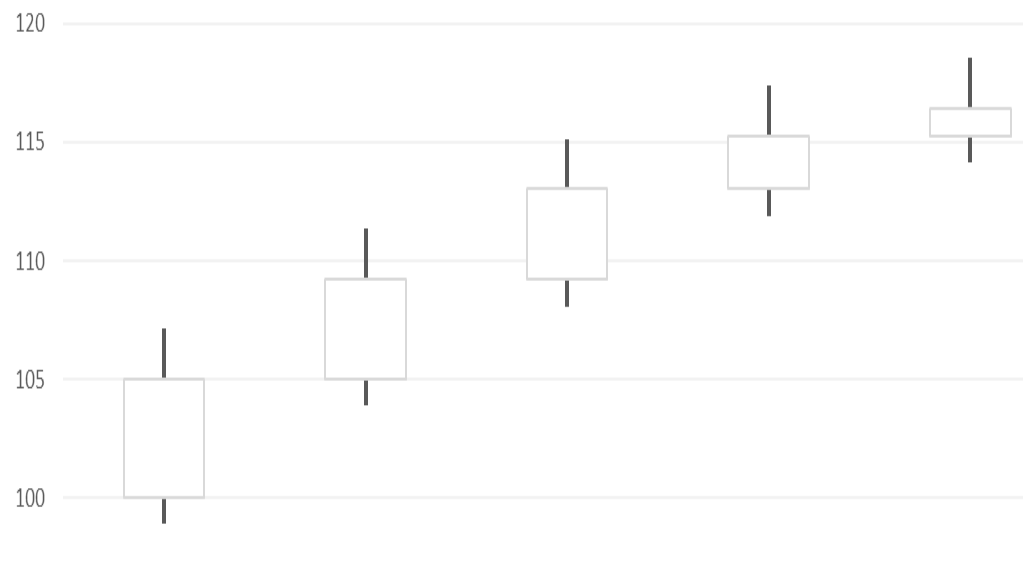}
		\caption{Ascending subseries}
		\label{fig:subfigA}
	\end{subfigure}
	\begin{subfigure}{0.45\linewidth}
		\includegraphics[width=\linewidth]{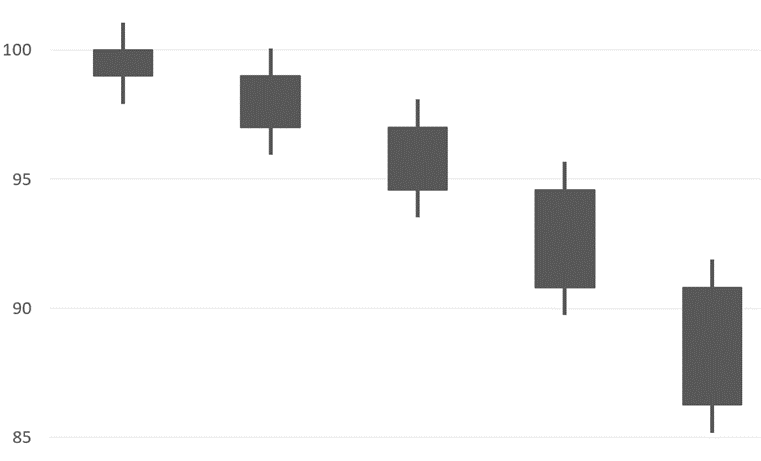}
		\caption{Descending subseries}
		\label{fig:subfigB}
	\end{subfigure}
	
	\caption{Candlestick chart of two subseries with the same acceleration.}
	\label{fig:Candlestick}
\end{figure}

If we consider the fluctuation (increase-decrease rate) of the points of a graph as important for training a model, we should refer to the price acceleration graph (price fluctuation fluctuation) from the price fluctuation graph. In this case, the two subseries of price fluctuations, which show the amount of price fluctuation at each moment compared to the price of the previous moment, if they have a similar shape in terms of appearance, they will have relatively the same acceleration graph. Even if, in terms of appearance, one subseries of fluctuations is above the zero line of the chart, and another subseries is exactly the same, but below the chart, they will have the same slope chart. Figure \ref{fig:enter-label} shows two subseries with the same slope, one completely above and one completely below the zero line. In fact, the two mentioned subseries show completely different price trends. In the first case, when the subseries is completely above the zero line, it indicates the upward trend of the price; in the second case, when the subseries is completely below the chart, it indicates the downward trend of the price. These two completely different states will likely show the same behavior in the next moment. To show this issue, you should refer to the candlestick chart of the initial time series. Figure \ref{fig:Candlestick} shows the candlestick chart of two subseries of Figure \ref{fig:enter-label}, where part A shows a completely bullish price and part B shows a completely bearish price.

One of the basic technical analyses in the candlestick structure is that if the increase in price gradually decreases, we will probably see a downward price trend in the future, and it can be a good point to sell that cryptocurrency. In another analysis, if the downward trend of the moment-to-moment price increases, the price of the next moment will probably decrease as well. Therefore, in both cases, which are completely different, there is a possibility that the next price will be downward. In Figure \ref{fig:enter-label} blue line, if we predict that the speed of price changes is downward, the degree of downward speed is very important. Because if the rate of change of the downward price and a number is higher than 0, the price will be upward. Also, if the rate of price change is downward and the number is smaller than 0, then the price will be bearish. In this example, The rate of change is decreasing, which has a great impact on investors's decision, will be predicted by the transformer-based model corresponding to the same category. In general, the model would predict the decrease of the fluctuation, but the amount of this fluctuation will be predicted according to the previous moments of the subseries.

\subsection{Probabilistic Model Selector}\label{subsec2}

We build a probability-based model with fast self-updating capability to predict the next subseries category and select its corresponding model. The model is made according to the history and possibilities of transfer from one category to another. If, when creating subseries from the main series, we move each time window stage by only one transfer step, and categorization according to the direction (up/down) of the price, in this model each category can only be transferred to two other categories. This way, there will be a possibility of transfer from each category to only two other categories.

\subsection{Temporal Fusion Transformer Construction}\label{subsec2tft}



The transformer model based on the attention mechanism was introduced by Vaswani et al. \cite{vaswani2017attention}. Attention models attempt to address the potential limitations of the encoder-decoder model and enable the network to focus on a few specific aspects at a time and ignore the rest. These models perform well in predicting time series due to paying attention to previous effective situations. Each time series record can be seen as an entity with three representation vectors: query vector, key vector, and value vector. The attention of an entity on a set of entities is a linear combination of the value vectors of the set, where the weights are determined by how similar the query vector of the entity is to the key vectors of each entity in the set. Cosine multiplication is used to measure the similarity, and then a softmax layer normalizes the similarity scores between zero and one.





Transformers are sequence-to-sequence (encoder-decoder) models that use the attention mechanism without using recursive mechanisms. The transformer encoder sub-network takes a text sequence as input and produces its representation. The decoder sub-network takes the representation of the encoder sub-network and the time series created so far (which are produced in the decoder) as input and tries to estimate the probability distribution of the next word. 
In the encoder part, each layer consists of two internal blocks, including the block of the self-attention mechanism and the fully connected forward neural network block. Residual connections and layer normalizer are used in each block. The decoding part is the same as the encoder, with a slight difference. The first difference is that each layer has three blocks. The first and third blocks of the decoder are the same as the first and second blocks of the encoder. The middle block connects the encoder and the decoder and operates attention on the output of the last layer of the encoder by means of a multi-head attention. Another difference between the decoder and the encoder is that in the first block (self-attention), each encoder token pays attention to all other tokens in the input sequence. However, in the first block of the decoder, each token of the destination sequence will only pay attention to the tokens before it.


Since the processing of inputs can be done in parallel, it is necessary to give the position of each input in the time series as an input to the network. After finishing the processes in the decoder module, the output is given to a feedforward layer with a smooth maximum activation function to predict the output string's next state (word). The number of neurons in this layer equals the number of states in the model. As mentioned before, the state generated at each time step, together with the previous states, is fed back to the decoder module to use them to generate the next state.

Recently, Lim et al. \cite{lim2021temporal} introduced a method called Temporal Fusion Transformer (TFT), that in addition to addressing the complexities present in recurrent neural networks like LSTM, takes covariates of each time frame, typically time-dependent, as inputs. The model incorporates attention to other previous time steps using the transformer approach, as proposed by Vaswani et al. \cite{vaswani2017attention}. Another feature in almost all TFT stages and gates is the ability to jump and direct connection, which makes it possible to skip stages and gates in situations where our data is simpler than the model. In this way, the model can adjust its complexity during training. Also, this model can receive auxiliary time-dependent variables in addition to the main time series and adjust the time in the time series according to the auxiliary variables.



If our data is less complex than the model, with the help of this model, it is possible to skip steps and gates. In this way, the model can adjust its complexity during training. The specialty of this model lies in its ability to observe a time series and predict its future continuation of the series. This model can receive time-dependent covariates in addition to the main time series and adjust the time in the time series according to the covariates. This feature of the model helps us to easily and without creating additional noise, connect the subseries of the same category that are separate and related to different times and give it as training data to the model.

In order to send the subseries of each category to the TFT model, we first join all the subseries of each category and generate a new time series. Then, we create a matrix with the length of the new time series, the value of each of which is equal to the index of the corresponding moment in the initial subseries. In this way, if each time subseries contains 7 moments, the constructed matrix contains the numbers 1, 2, 3, ..., 7 that are repeated alternately. This matrix will be sent as time-dependent covariates data to the TFT model. The mentioned matrix is one-dimensional, which can be converted into two-dimensional, and other time parameters, such as hour, day of the year, day of the week, etc., from the primary subseries can be placed in the other dimension of the matrix.

After creating the time series of each category and the corresponding covariates data matrix, we will send them to the TFT models for training. This way, we will have a TFT model corresponding to each category. After training the models, to predict the new data, we must first find the category of the last subseries of the new data. Then, with the help of a probabilistic selector model, we predict the next subseries category. Therefore, to predict the price of the next moment, it is enough to send the next subseries to the TFT model of the predicted category.

\section{Experiments and Results}\label{sec2}

In this section, we study influential features for cryptocurrency price prediction and evaluate the performance of the proposed model against other methods. 

\subsection{Dataset}\label{datasetsubsec2}

A large amount of cryptocurrency transfers happen through decentralized exchanges, which were used in this research. All transactions are entered in currency pairs, i.e., the price of the first cryptocurrency compared to the second. We store transaction information of Binance exchange as the largest decentralized cryptocurrency exchange cumulatively in 1-minute time intervals for 18 months since December 2021. Therefore, each database record represents the transactions of a currency pair in a specific minute \cite{githubGitHubArashitc2Binance1minutecandles}. Due to the unavailability of one-minute transaction details for an extended period in the existing datasets and APIs, we had to gather the necessary information.

All buying and selling orders are divided into two categories: taker and maker. If a cryptocurrency buyer buys instantaneously at the price listed by the sellers, it is called a \textit{taker}. The Taker buyer chooses from the sale list and makes his purchase instantly. However, if the buyer wants to buy at a price lower than the selling list, he will register his order in the buying queue so that whenever a seller agrees to that price, the transaction will be done. In this case, the buyer is called a \textit{maker}. The Maker buyer does not make his purchase instantly and registers his order in the shopping queue. In the same way, the sellers are also divided into two categories: Takers and Makers. 

The information stored in each record includes the time of the start and end of that minute in the form of a timestamp, the title of the cryptocurrency pair, the value of the first cryptocurrency compared to the second cryptocurrency at the exact start time (the first transaction of that minute), the value of the first cryptocurrency compared to the second one at the end of the minute (the last transaction of that minute), the maximum and minimum value of the first cryptocurrency compared to the second one in that minute, the trading volume of that minute based on the first cryptocurrency (V), the trading volume based on the second cryptocurrency (QAV), the number of transactions, the volume purchased by the taker buyer, the payment amount of the second cryptocurrency to buy the first cryptocurrency by the taker buyer, and the name of the exchange. All the mentioned information has been received from Binance decentralized exchange, which is known as the largest and most effective decentralized cryptocurrency exchange. We examined 13 currency pairs for this research and collected in our database. The first cryptocurrencies of these currency pairs include Bitcoin, Ethereum, Bitcoin Cash, Litecoin, Dogecoin, PolkaDOT, and Ripple.


\subsection{Implementation and Settings}\label{subsec2}
We used Python \textit{Darts} library to implement the main predictive models of TFT. We also implemented and tested the TFT Model Selector in different ways including Markov chain and LSTM model using \textit{Keras} library. Our code and tests are publicly available on Github \cite{githubGitHubArashitc2sourcecode}.

\textbf{Time frame.} We initially conducted our experiment on 1-minute time frames; however, since the time frame is very small and at any moment the minimum price of the selling queue is different from the maximum price of the buying queue and the last transaction of the exchange may be done from one of these two queues, therefore, small fluctuations are created at any moment. Transaction information of these moments and fluctuations alone are not meaningful, and by aggregating a volume of these moments, we will get meaningful information. After conducting experiments, we empirically concluded that the minimum meaningful small time frame is 7 minutes. 

\textbf{TFT models.} Temporal Fusion Transformers are powerful models for predicting multivariate time series \cite{lim2021temporal}. In our proposed method, we use a TFT model to predict each category of subseries to predict the next moment price change of those subseries with higher accuracy. The TFT gets as input static metadata, time-varying past inputs, and time-varying a priori known future inputs. Given time subseries in 8 time frames, we want to predict the price or direction of the closing moment of the 8th time frame by observing 7 time frames of 7 minutes. Therefore, in the created TFT models, we set the value of input-chunk-length to 7 and the value of output-chunk-length to 1. We also set the Hidden state size of the TFT to 70, used 4 LSTM layers and 4 attention heads, and trained the model with only 7 epochs. According to the number of 8 time frames in the existing time subseries, if we consider the upward or downward conditions of the close price of each time frame compared to the previous time frame, we will have \(2^7\) different subseries. We created a TFT model for each state of the subseries. We further created a Markov chain-type model that has the status of the number of TFT models made (128 models). From each situation, there is a possibility of moving to only two other situations (upward or downward price at the close of the next time frame). This model is responsible for selecting the appropriate TFT model according to the probabilities to predict the 8th price direction. For testing purposes, instead of 128 TFT models produced and the Markovi selector model, TFT models were created that did not use the last minute time frame in model making. These models predict the last moment situation (upward or downward) without using a selector. Therefore, in this experiment, \(2^6\) classes of subseries were created, equivalent to 64 different TFT models.


The TFT model can learn the pattern well by receiving covariates that show a pattern in the time series. Since the TFT model receives a continuous time series for training, we concatenate them to send the subseries of a category to the model. In order for the model to understand and focus on the repeatability of the subseries behavior, we assigned a sequence number to each moment of each subseries and sent them to the model as covariates.
In these models, we used 4 layers for the LSTM Encoder and Decoder. Also, we used 4 attention heads in making the model.

\textbf{TFT model selector.} The selector decides which TFT model corresponding to the category can better predict the price volatility rate. We evaluate 3 different solutions for this selector. First, we create a Markov chain to predict the next category that will be observed according to the probabilities (HMM). Second, we use LSTM model to predict the next category. After predicting the next category, we will send the last sub-time series to the corresponding model of that category to predict the price situation of the next moment. Third, we delegate the prediction of the next category to the same TFT model. In this way, from the same stage of categorization, we do not enter the moment of the tag in the categorization and let our TFT model recognize that the price volatility rate of the final moment has decreased or increased compared to the previous moment. In this case, the number of moments based on which the categorization is done will be reduced by one moment, and as a result, the number of categories will be halved. Note that in the first and the second method, i.e., HMM and LSTM, we predict the increase or decrease of the price volatility rate compared to the previous moment with the selector model. However, we use the TFT model to predict whether this increase or decrease in speed will cause the volatility rate to become positive or negative, i.e., increase or decrease in the price of the cryptocurrency.

\subsection{Comparison}\label{subsec2}

\textbf{Performance.} We assess our model's prediction accuracy and precision compared to the baselines. For testing, we used LTC-USDT transaction data that were not present in the model training from 08-22-2023 for two weeks. We checked the criteria only for bullish or bearish prices. In trading, if a bearish time frame is wrongly recognized as bullish due to the purchase of cryptocurrency and entry into the trade, in addition to not achieving the estimated profit, the previous asset will also be damaged. For this reason, the precision of the methods will be calculated and evaluated. On the other hand, if the bullish time frame is wrongly recognized as bearish, the previous asset will not suffer a loss. Table \ref{tab-Accuracy-Precision} shows the accuracy and precision of our proposed method compared to recent basic techniques. 

In this comparison, we did not include financial covariates in any of the TFT models. Results do not show a high value for accuracy and precision numbers and also the improvement does not seem to be considerable. However, note that in this work in order to avoid very large amount of transactions we confined our time frame to 7 minutes that is very small compared to hourly or daily time frames. Cryptocurrency markets are chaotic in small time frames and due to the large number of transactions any small improvement in accuracy and precision values can have a great impact on long-term profits, e.g. on a monthly basis. In general, we are focused on short time frames and quick profitable trades. This strategy is designed to take advantage of short-term market movements by analyzing price action on a short time frames, and is known for its agility and ability to capture quick profits in volatile markets.

\begin{table}[h]
\caption{Accuracy and Precision Comparison.}\label{tab-Accuracy-Precision}%
\begin{tabular}{lll}
\toprule
Model & Accuracy & Precision \\
\midrule
Trading using LSTM model \cite{song2020time}   & 0.496 & 0.492  \\
Trading using a TFT model \cite{lim2021temporal} without financial covariates    & 0.535 & 0.529  \\
Trading with subseries categorization   & \textbf{0.539} & \textbf{0.534} \\
(without selector model and 64 categories)\footnotemark[1]   & \\

\botrule
\end{tabular}
\footnotetext[1]{Our proposed method in simple mode}
\end{table}

\textbf{Selector accuracy.} In two of our experiments, we used a separate selector model including Markov chain and LSTM, and in another experiment, we combined the selector with the original TFT models. For LTC-USDT currency pair, the accuracy of TFT model selector was 0.79 for Markov chain selector and 0.74 for LSTM selector.


\textbf{Profitability.} We calculate the profitability through a simulation. We use the dataset from the previous section for testing. In this experiment, we simulate a space in which we have 100 USDT at the beginning of the asset period, and with the help of the proposed methods, based on the models' output buy and sell signals, we trade at that moment's price. In these transactions, we considered the 0 exchange fee. We entered the entire asset into the transaction with each signal, regardless of the transaction's risk level and the stock portfolio's management. The end-of-period asset represents a measure of the profitability of the proposed method. Table ~\ref{tab-profitability} shows the profitability of the proposed methods in this simulation. Note that a 1 USDT increase in profit in 2 weeks based on the initial balance of 100 USDT is equivalent to an increase of 2\% in monthly profit compared to other methods, and this profit is achieved in a bear market without using futures and leverages. In our analysis time frame, without trading the 100 USDT value decreases to 98 and using LSTM model for training it gets even less to 97.55. Using other baselines will almost retain 100 USDT as shown in Table 2, except for trading using a TFT model using financial covariates that is comparable to ours, while in our experiments we did not consider financial covariates and left it for a future work. The 4 USDT increase means 4\% profit in 2 weeks that is about 8\% increase per month that is a considerable profit.

\begin{table}[h]
\caption{Profitability Comparison.}\label{tab-profitability}%
\begin{tabular}{ll}
\toprule
Model & End of term\\
 & asset amount\\
\midrule
Without trading (buying at the beginning and selling at the end of the period) & 98.01 USDT   \\
Trading using LSTM model   & 97.55 USDT \\
Forecasting and trading using a TFT model without financial covariates\footnotemark[1]    & 100.81 USDT \\
Forecasting and trading using a TFT model using financial covariates   & 103.19 USDT \\
Trading by categorization of subseries and using LSTM selector model   & 100.35 USDT\\
Trading by categorization of subseries and using Markov chain based selector    & 103.16 USDT\\
Trading with subseries categorization (without selector model and 64 categories)   & \textbf{104.08} USDT\\

\botrule
\end{tabular}
\footnotetext{Source: The profitability of the model in Litecoin trading in a two-week period from 08/22/2023.}
\footnotetext[1]{Similar conditions to the proposed model.}
\end{table}

\section{Discussion}\label{sec2}
\textbf{Profitability and accuracy.} We observed in our experiments with the dataset for testing that the price of the selected cryptocurrency is generally down 2\% in the selected 14 days. Using our proposed model we achieved more than 4\% profit, which compared to the profitability of holding cryptocurrency in this period, our temporal fusion transformer models with time series categorization have created more than 6\% more profit in 2 weeks. In the same conditions, the proposed model performs better than LSTM models and transformers (without classification), while only using price data and not any other covariates.

We also evaluated different selectors by using TFT models corresponding to each category. First, in order to evaluate the accuracy of the category models, we assumed a selector that selects the category (prediction model) of each subseries with 100\% accuracy. In this case, the category models turn 100 USDT into 248 USDT in two weeks, and actually make a 148\% profit in two weeks. Then we created a selector with LSTM, but contrary to expectations it reduced the amount of assets and after two weeks \$2.50 of the initial assets were gone. Another selector that was tested was based on probabilities and Markov chain. This model had more than 3\% profitability in the mentioned two weeks. The important feature of this selector is that it can be used in predicting other cryptocurrencies without changing the transformer-based models and only by changing the probability values of the Markov chain. We also tested a method without using a selector so that our trained TFT models also predict the bullishness or the bearishness of the price velocity. When categorizing the subseries, we do not use the price fluctuation of the last time frame in the categorization, and the model should predict the entire situation of the last moment. In this method, due to the reduction of a time frame in categorization, we have 64 categories, and the training data of each category increases. We obtained more than 4\% profit within two weeks of testing using this approach. Here the disadvantage is that the general models may show different qualities and accuracy in other cryptocurrencies.

\textbf{Generalizability.} With regards to the generalizability of our proposed approach on other financial markets, such as Foreign Exchange (Forex) or commodities like gold, or stock market, we can similarly utilize trading data of such markets. Our method does not depend on only one specific cryptocurrency due to the combination of sub-series of different cryptocurrencies in the construction of TFT models, and the models learn the behavior of a category of sub-series. Therefore, the learned model can be used in other financial markets by creating the TFT Model selector for that financial market. Also, according to the previous work done in fractal dimensions \cite{bhatt2015fractal}, with some changes, the data of other time frames can be predicted with this built model and we leave this as a future work. The produced TFT models can be affected by the length of time frames and in financial markets with extreme fluctuations, if the length of the time frame increases, the behaviors may intensify and the TFT models may not work well.

\section{Conclusion and Future Work}\label{sec2}

In this paper, we presented a method for predicting financial markets that lack extensive historical data. By using statistical models such as the hidden Markov model for the selector, we can update the model in real-time. This allows us to observe the varying behaviors of time series dynamically and simultaneously enhance the model's training. Our results indicate that if the category selector model achieved 100\% accuracy, a profit of 148\% could be obtained. However, with a selector accuracy of 79\%, we achieved a profit of 3.16\% over a two-week period. Therefore, even minor improvements in the selector's accuracy can significantly increase profitability. Enhancing the accuracy of the selector model will be a primary focus of our future work.

Our experiments show that the profitability of the model built using our proposed method, without utilizing any financial covariates, is higher than that of a simple TFT model that does use covariates. We will cconsider incorporating covariates into the construction of TFT models or the selector model as a future research to further enhance model accuracy and profitability. This work contributes to the state-of-the-art by offering an adaptive method for financial market prediction. The importance of our work lies in its potential application across various financial markets, providing a flexible and continually improving predictive model.

\bibliography{sn-bibliography}


\begin{thebibliography}{26}
\ifx \bisbn   \undefined \def \bisbn  #1{ISBN #1}\fi
\ifx \binits  \undefined \def \binits#1{#1}\fi
\ifx \bauthor  \undefined \def \bauthor#1{#1}\fi
\ifx \batitle  \undefined \def \batitle#1{#1}\fi
\ifx \bjtitle  \undefined \def \bjtitle#1{#1}\fi
\ifx \bvolume  \undefined \def \bvolume#1{\textbf{#1}}\fi
\ifx \byear  \undefined \def \byear#1{#1}\fi
\ifx \bissue  \undefined \def \bissue#1{#1}\fi
\ifx \bfpage  \undefined \def \bfpage#1{#1}\fi
\ifx \blpage  \undefined \def \blpage #1{#1}\fi
\ifx \burl  \undefined \def \burl#1{\textsf{#1}}\fi
\ifx \doiurl  \undefined \def \doiurl#1{\url{https://doi.org/#1}}\fi
\ifx \betal  \undefined \def \betal{\textit{et al.}}\fi
\ifx \binstitute  \undefined \def \binstitute#1{#1}\fi
\ifx \binstitutionaled  \undefined \def \binstitutionaled#1{#1}\fi
\ifx \bctitle  \undefined \def \bctitle#1{#1}\fi
\ifx \beditor  \undefined \def \beditor#1{#1}\fi
\ifx \bpublisher  \undefined \def \bpublisher#1{#1}\fi
\ifx \bbtitle  \undefined \def \bbtitle#1{#1}\fi
\ifx \bedition  \undefined \def \bedition#1{#1}\fi
\ifx \bseriesno  \undefined \def \bseriesno#1{#1}\fi
\ifx \blocation  \undefined \def \blocation#1{#1}\fi
\ifx \bsertitle  \undefined \def \bsertitle#1{#1}\fi
\ifx \bsnm \undefined \def \bsnm#1{#1}\fi
\ifx \bsuffix \undefined \def \bsuffix#1{#1}\fi
\ifx \bparticle \undefined \def \bparticle#1{#1}\fi
\ifx \barticle \undefined \def \barticle#1{#1}\fi
\bibcommenthead
\ifx \bconfdate \undefined \def \bconfdate #1{#1}\fi
\ifx \botherref \undefined \def \botherref #1{#1}\fi
\ifx \url \undefined \def \url#1{\textsf{#1}}\fi
\ifx \bchapter \undefined \def \bchapter#1{#1}\fi
\ifx \bbook \undefined \def \bbook#1{#1}\fi
\ifx \bcomment \undefined \def \bcomment#1{#1}\fi
\ifx \oauthor \undefined \def \oauthor#1{#1}\fi
\ifx \citeauthoryear \undefined \def \citeauthoryear#1{#1}\fi
\ifx \endbibitem  \undefined \def \endbibitem {}\fi
\ifx \bconflocation  \undefined \def \bconflocation#1{#1}\fi
\ifx \arxivurl  \undefined \def \arxivurl#1{\textsf{#1}}\fi
\csname PreBibitemsHook\endcsname

\bibitem[\protect\citeauthoryear{Corbet
  et~al.}{2019}]{corbet2019cryptocurrencies}
\begin{barticle}
\bauthor{\bsnm{Corbet}, \binits{S.}},
\bauthor{\bsnm{Lucey}, \binits{B.}},
\bauthor{\bsnm{Urquhart}, \binits{A.}},
\bauthor{\bsnm{Yarovaya}, \binits{L.}}:
\batitle{Cryptocurrencies as a financial asset: A systematic analysis}.
\bjtitle{International Review of Financial Analysis}
\bvolume{62},
\bfpage{182}--\blpage{199}
(\byear{2019})
\end{barticle}
\endbibitem

\bibitem[\protect\citeauthoryear{Angela and Sun}{2020}]{angela2020factors}
\begin{bchapter}
\bauthor{\bsnm{Angela}, \binits{O.}},
\bauthor{\bsnm{Sun}, \binits{Y.}}:
\bctitle{Factors affecting cryptocurrency prices: Evidence from ethereum}.
In: \bbtitle{2020 International Conference on Information Management and
  Technology (ICIMTech)},
pp. \bfpage{318}--\blpage{323}
(\byear{2020}).
\bcomment{IEEE}
\end{bchapter}
\endbibitem

\bibitem[\protect\citeauthoryear{Sovbetov}{2018}]{sovbetov2018factors}
\begin{barticle}
\bauthor{\bsnm{Sovbetov}, \binits{Y.}}:
\batitle{Factors influencing cryptocurrency prices: Evidence from bitcoin,
  ethereum, dash, litcoin, and monero}.
\bjtitle{Journal of Economics and Financial Analysis}
\bvolume{2}(\bissue{2}),
\bfpage{1}--\blpage{27}
(\byear{2018})
\end{barticle}
\endbibitem

\bibitem[\protect\citeauthoryear{Yamin and
  Chaudhry}{2023}]{yamin2023cryptocurrency}
\begin{botherref}
\oauthor{\bsnm{Yamin}, \binits{M.A.}},
\oauthor{\bsnm{Chaudhry}, \binits{M.}}:
Cryptocurrency market trend and direction prediction using machine learning: A
  comprehensive survey
(2023)
\end{botherref}
\endbibitem

\bibitem[\protect\citeauthoryear{Akyildirim
  et~al.}{2023}]{akyildirim2023forecasting}
\begin{barticle}
\bauthor{\bsnm{Akyildirim}, \binits{E.}},
\bauthor{\bsnm{Cepni}, \binits{O.}},
\bauthor{\bsnm{Corbet}, \binits{S.}},
\bauthor{\bsnm{Uddin}, \binits{G.S.}}:
\batitle{Forecasting mid-price movement of bitcoin futures using machine
  learning}.
\bjtitle{Annals of Operations Research}
\bvolume{330}(\bissue{1}),
\bfpage{553}--\blpage{584}
(\byear{2023})
\end{barticle}
\endbibitem

\bibitem[\protect\citeauthoryear{Chen et~al.}{2019}]{chen2019predicting}
\begin{botherref}
\oauthor{\bsnm{Chen}, \binits{M.}},
\oauthor{\bsnm{Narwal}, \binits{N.}},
\oauthor{\bsnm{Schultz}, \binits{M.}}:
Predicting price changes in ethereum.
International Journal on Computer Science and Engineering (IJCSE) ISSN,
0975--3397
(2019)
\end{botherref}
\endbibitem

\bibitem[\protect\citeauthoryear{Fazlollahi and
  Ebrahimijam}{2023}]{fazlollahi2023predicting}
\begin{bchapter}
\bauthor{\bsnm{Fazlollahi}, \binits{N.}},
\bauthor{\bsnm{Ebrahimijam}, \binits{S.}}:
\bctitle{Predicting cryptocurrency price returns by using deep learning model
  of technical analysis indicators}.
In: \bbtitle{Global Economic Challenges: 6th International Conference on
  Banking and Finance Perspectives, Cuenca, Spain},
pp. \bfpage{175}--\blpage{186}
(\byear{2023}).
\bcomment{Springer}
\end{bchapter}
\endbibitem

\bibitem[\protect\citeauthoryear{Bouteska
  et~al.}{2024}]{bouteska2024cryptocurrency}
\begin{barticle}
\bauthor{\bsnm{Bouteska}, \binits{A.}},
\bauthor{\bsnm{Abedin}, \binits{M.Z.}},
\bauthor{\bsnm{Hajek}, \binits{P.}},
\bauthor{\bsnm{Yuan}, \binits{K.}}:
\batitle{Cryptocurrency price forecasting--a comparative analysis of ensemble
  learning and deep learning methods}.
\bjtitle{International Review of Financial Analysis}
\bvolume{92},
\bfpage{103055}
(\byear{2024})
\end{barticle}
\endbibitem

\bibitem[\protect\citeauthoryear{Lim et~al.}{2021}]{lim2021temporal}
\begin{barticle}
\bauthor{\bsnm{Lim}, \binits{B.}},
\bauthor{\bsnm{Ar{\i}k}, \binits{S.{\"O}.}},
\bauthor{\bsnm{Loeff}, \binits{N.}},
\bauthor{\bsnm{Pfister}, \binits{T.}}:
\batitle{Temporal fusion transformers for interpretable multi-horizon time
  series forecasting}.
\bjtitle{International Journal of Forecasting}
\bvolume{37}(\bissue{4}),
\bfpage{1748}--\blpage{1764}
(\byear{2021})
\end{barticle}
\endbibitem

\bibitem[\protect\citeauthoryear{Vaswani et~al.}{2017}]{vaswani2017attention}
\begin{botherref}
\oauthor{\bsnm{Vaswani}, \binits{A.}},
\oauthor{\bsnm{Shazeer}, \binits{N.}},
\oauthor{\bsnm{Parmar}, \binits{N.}},
\oauthor{\bsnm{Uszkoreit}, \binits{J.}},
\oauthor{\bsnm{Jones}, \binits{L.}},
\oauthor{\bsnm{Gomez}, \binits{A.N.}},
\oauthor{\bsnm{Kaiser}, \binits{{\L}.}},
\oauthor{\bsnm{Polosukhin}, \binits{I.}}:
Attention is all you need.
Advances in neural information processing systems
\textbf{30}
(2017)
\end{botherref}
\endbibitem

\bibitem[\protect\citeauthoryear{Hu}{2021}]{hu2021stock}
\begin{bchapter}
\bauthor{\bsnm{Hu}, \binits{X.}}:
\bctitle{Stock price prediction based on temporal fusion transformer}.
In: \bbtitle{2021 3rd International Conference on Machine Learning, Big Data
  and Business Intelligence (MLBDBI)},
pp. \bfpage{60}--\blpage{66}
(\byear{2021}).
\bcomment{IEEE}
\end{bchapter}
\endbibitem

\bibitem[\protect\citeauthoryear{Kumar et~al.}{2018}]{kumar2018comparative}
\begin{bchapter}
\bauthor{\bsnm{Kumar}, \binits{I.}},
\bauthor{\bsnm{Dogra}, \binits{K.}},
\bauthor{\bsnm{Utreja}, \binits{C.}},
\bauthor{\bsnm{Yadav}, \binits{P.}}:
\bctitle{A comparative study of supervised machine learning algorithms for
  stock market trend prediction}.
In: \bbtitle{2018 Second International Conference on Inventive Communication
  and Computational Technologies (ICICCT)},
pp. \bfpage{1003}--\blpage{1007}
(\byear{2018}).
\bcomment{IEEE}
\end{bchapter}
\endbibitem

\bibitem[\protect\citeauthoryear{Song and Lee}{2019}]{song2019design}
\begin{bchapter}
\bauthor{\bsnm{Song}, \binits{Y.}},
\bauthor{\bsnm{Lee}, \binits{J.}}:
\bctitle{Design of stock price prediction model with various configuration of
  input features}.
In: \bbtitle{Proceedings of the International Conference on Artificial
  Intelligence, Information Processing and Cloud Computing},
pp. \bfpage{1}--\blpage{5}
(\byear{2019})
\end{bchapter}
\endbibitem

\bibitem[\protect\citeauthoryear{Werawithayaset and
  Tritilanunt}{2019}]{werawithayaset2019stock}
\begin{bchapter}
\bauthor{\bsnm{Werawithayaset}, \binits{P.}},
\bauthor{\bsnm{Tritilanunt}, \binits{S.}}:
\bctitle{Stock closing price prediction using machine learning}.
In: \bbtitle{2019 17th International Conference on ICT and Knowledge
  Engineering (ICT\&KE)},
pp. \bfpage{1}--\blpage{8}
(\byear{2019}).
\bcomment{IEEE}
\end{bchapter}
\endbibitem

\bibitem[\protect\citeauthoryear{Xingzhou
  et~al.}{2019}]{xingzhou2019predictive}
\begin{bchapter}
\bauthor{\bsnm{Xingzhou}, \binits{L.}},
\bauthor{\bsnm{Hong}, \binits{R.}},
\bauthor{\bsnm{Yujun}, \binits{Z.}}:
\bctitle{Predictive modeling of stock indexes using machine learning and
  information theory}.
In: \bbtitle{Proceedings of the 2019 10th International Conference on
  E-business, Management and Economics},
pp. \bfpage{175}--\blpage{179}
(\byear{2019})
\end{bchapter}
\endbibitem

\bibitem[\protect\citeauthoryear{Sarode et~al.}{2019}]{sarode2019stock}
\begin{bchapter}
\bauthor{\bsnm{Sarode}, \binits{S.}},
\bauthor{\bsnm{Tolani}, \binits{H.G.}},
\bauthor{\bsnm{Kak}, \binits{P.}},
\bauthor{\bsnm{Lifna}, \binits{C.}}:
\bctitle{Stock price prediction using machine learning techniques}.
In: \bbtitle{2019 International Conference on Intelligent Sustainable Systems
  (ICISS)},
pp. \bfpage{177}--\blpage{181}
(\byear{2019}).
\bcomment{IEEE}
\end{bchapter}
\endbibitem

\bibitem[\protect\citeauthoryear{Ingle and Deshmukh}{2016}]{ingle2016hidden}
\begin{bchapter}
\bauthor{\bsnm{Ingle}, \binits{V.}},
\bauthor{\bsnm{Deshmukh}, \binits{S.}}:
\bctitle{Hidden markov model implementation for prediction of stock prices with
  tf-idf features}.
In: \bbtitle{Proceedings of the International Conference on Advances in
  Information Communication Technology \& Computing},
pp. \bfpage{1}--\blpage{6}
(\byear{2016})
\end{bchapter}
\endbibitem

\bibitem[\protect\citeauthoryear{Singh et~al.}{2019}]{singh2019stock}
\begin{bchapter}
\bauthor{\bsnm{Singh}, \binits{S.}},
\bauthor{\bsnm{Madan}, \binits{T.K.}},
\bauthor{\bsnm{Kumar}, \binits{J.}},
\bauthor{\bsnm{Singh}, \binits{A.K.}}:
\bctitle{Stock market forecasting using machine learning: Today and tomorrow}.
In: \bbtitle{2019 2nd International Conference on Intelligent Computing,
  Instrumentation and Control Technologies (ICICICT)},
vol. \bseriesno{1},
pp. \bfpage{738}--\blpage{745}
(\byear{2019}).
\bcomment{IEEE}
\end{bchapter}
\endbibitem

\bibitem[\protect\citeauthoryear{Ramos-P{\'e}rez et~al.}{2021}]{ramos2021multi}
\begin{barticle}
\bauthor{\bsnm{Ramos-P{\'e}rez}, \binits{E.}},
\bauthor{\bsnm{Alonso-Gonz{\'a}lez}, \binits{P.J.}},
\bauthor{\bsnm{N{\'u}{\~n}ez-Vel{\'a}zquez}, \binits{J.J.}}:
\batitle{Multi-transformer: A new neural network-based architecture for
  forecasting s\&p volatility}.
\bjtitle{Mathematics}
\bvolume{9}(\bissue{15}),
\bfpage{1794}
(\byear{2021})
\end{barticle}
\endbibitem

\bibitem[\protect\citeauthoryear{Anbaee~Farimani
  et~al.}{2021}]{farimani2021leveraging}
\begin{bchapter}
\bauthor{\bsnm{Anbaee~Farimani}, \binits{S.}},
\bauthor{\bsnm{Vafaei~Jahan}, \binits{M.}},
\bauthor{\bsnm{Milani~Fard}, \binits{A.}},
\bauthor{\bsnm{Haffari}, \binits{G.}}:
\bctitle{Leveraging latent economic concepts and sentiments in the news for
  market prediction}.
In: \bbtitle{2021 IEEE 8th International Conference on Data Science and
  Advanced Analytics (DSAA)},
pp. \bfpage{1}--\blpage{10}
(\byear{2021}).
\bcomment{IEEE}
\end{bchapter}
\endbibitem

\bibitem[\protect\citeauthoryear{Anbaee~Farimani
  et~al.}{2022}]{farimani2022investigating}
\begin{barticle}
\bauthor{\bsnm{Anbaee~Farimani}, \binits{S.}},
\bauthor{\bsnm{Vafaei~Jahan}, \binits{M.}},
\bauthor{\bsnm{Milani~Fard}, \binits{A.}},
\bauthor{\bsnm{Tabbakh}, \binits{S.R.K.}}:
\batitle{Investigating the informativeness of technical indicators and news
  sentiment in financial market price prediction}.
\bjtitle{Knowledge-Based Systems}
\bvolume{247},
\bfpage{108742}
(\byear{2022})
\end{barticle}
\endbibitem

\bibitem[\protect\citeauthoryear{Anbaee~Farimani
  et~al.}{2024}]{farimani2024adaptive}
\begin{barticle}
\bauthor{\bsnm{Anbaee~Farimani}, \binits{S.}},
\bauthor{\bsnm{Vafaei~Jahan}, \binits{M.}},
\bauthor{\bsnm{Milani~Fard}, \binits{A.}}:
\batitle{An adaptive multimodal learning model for financial market price
  prediction}.
\bjtitle{IEEE Access}
\bvolume{12},
\bfpage{121846}--\blpage{121863}
(\byear{2024})
\doiurl{10.1109/ACCESS.2024.3441029}
\end{barticle}
\endbibitem

\bibitem[\protect\citeauthoryear{}{}]{githubGitHubArashitc2Binance1minutecandles}
\begin{botherref}
{G}it{H}ub - arashitc2/{B}inance-1-minute-candles.
\url{https://github.com/arashitc2/Binance-1-minute-candles}.
[Accessed 09-Jun-2023]
\end{botherref}
\endbibitem

\bibitem[\protect\citeauthoryear{}{}]{githubGitHubArashitc2sourcecode}
\begin{botherref}
{G}it{H}ub -
  arashitc2/{I}mproving-cryptocurrency-price-forecasting-using-time-series-categorization-and-tft.
\url{https://github.com/arashitc2/Improving-cryptocurrency-price-forecasting-using-time-series-categorization-and-tft}.
[Accessed 25-May-2024]
\end{botherref}
\endbibitem

\bibitem[\protect\citeauthoryear{Song et~al.}{2020}]{song2020time}
\begin{barticle}
\bauthor{\bsnm{Song}, \binits{X.}},
\bauthor{\bsnm{Liu}, \binits{Y.}},
\bauthor{\bsnm{Xue}, \binits{L.}},
\bauthor{\bsnm{Wang}, \binits{J.}},
\bauthor{\bsnm{Zhang}, \binits{J.}},
\bauthor{\bsnm{Wang}, \binits{J.}},
\bauthor{\bsnm{Jiang}, \binits{L.}},
\bauthor{\bsnm{Cheng}, \binits{Z.}}:
\batitle{Time-series well performance prediction based on long short-term
  memory (lstm) neural network model}.
\bjtitle{Journal of Petroleum Science and Engineering}
\bvolume{186},
\bfpage{106682}
(\byear{2020})
\end{barticle}
\endbibitem

\bibitem[\protect\citeauthoryear{Bhatt et~al.}{2015}]{bhatt2015fractal}
\begin{barticle}
\bauthor{\bsnm{Bhatt}, \binits{S.}},
\bauthor{\bsnm{Dedania}, \binits{H.}},
\bauthor{\bsnm{Shah}, \binits{V.R.}}:
\batitle{Fractal dimensional analysis in financial time series}.
\bjtitle{International Journal of Financial Management}
\bvolume{5}(\bissue{2}),
\bfpage{57}--\blpage{62}
(\byear{2015})
\end{barticle}
\endbibitem

\end{thebibliography}

\end{document}